\documentclass{sig-alternate-05-2015}

\usepackage{array,booktabs,makecell}
\usepackage{multirow}
\usepackage{amsmath}
\usepackage{bm}
\usepackage[linesnumbered,ruled]{algorithm2e}
\usepackage{mathtools}
\usepackage{dblfloatfix}
\usepackage{subfigure}
\usepackage[export]{adjustbox}
\usepackage{comment}
\usepackage{etoolbox}
\usepackage{enumerate}
\usepackage[pagebackref=true,breaklinks=true,letterpaper=true,colorlinks,bookmarks=false]{hyperref}
\newbool{inccomment}
\setbool{inccomment}{true}
\newcommand{\cc}[1]{\ifbool{inccomment}{{\color{blue}#1}}{}}
\newcommand{\fq}[1]{\ifbool{inccomment}{{\color{magenta}#1}}{}}

\begin{document}

\CopyrightYear{2018} 
\setcopyright{acmcopyright} 
\conferenceinfo{DAC '18,}{June 24--29, 2018, San Francisco, CA, USA}
\isbn{978-1-4503-5700-5/18/06}
\acmPrice{\$15.00}
\doi{https://doi.org/10.1145/3195970.3196131}

\title{Towards Accurate and High-Speed Spiking Neuromorphic Systems with Data Quantization-Aware Deep Networks}

\author{
	Fuqiang Liu and Chenchen Liu\\
	\email{fqliu92@gmail.com, chchenliu@gmail.com}
	}


\maketitle




\begin{abstract}

Deep Neural Networks (DNNs) have gained immense success in cognitive applications and greatly pushed today's artificial intelligence forward. 
The biggest challenge in executing DNNs is their extremely data-extensive computations.
The computing efficiency in speed and energy is constrained when traditional computing platforms are employed in such computational hungry executions.
Spiking neuromorphic computing (SNC) has been widely investigated in deep networks implementation own to their high efficiency in computation and communication. 
However, weights and signals of DNNs are required to be quantized when deploying the DNNs on the SNC, which results in unacceptable accuracy loss.
Previous works mainly focus on weights discretize while inter-layer signals are mainly neglected. 
In this work, we propose to represent DNNs with fixed integer inter-layer signals and fixed-point weights while holding good accuracy. 
We implement the proposed DNNs on the memristor-based SNC system as a deployment example.
With 4-bit data representation, our results show that the accuracy loss can be controlled within 0.02\% (2.3\%) on MNIST (CIFAR-10).
Compared with the 8-bit dynamic fixed-point DNNs, our system can achieve more than 9.8$\times$ speedup, 89.1\% energy saving, and 30\% area saving. 

\end{abstract}

\section{Introduction}

Deep Neural Networks (DNNs) have achieved great success in cognitive applications such as image classification~\cite{NIPS2012_4824,VGG,Resnet}, object detection~\cite{girshick14CVPR,renNIPS15fasterrcnn}, and natural language processing~\cite{Recent}. 
However, the computations are extremely data-extensive and expensive in perspective of speed and energy.
And the computing power of the current von Neumann machines with limited data bandwidth and energy efficiency becomes insufficient to support these computations.
This issue becomes more severe with the rapid growth of the depth of the deep network models~\cite{rastegariECCV16}.
Consequently, novel non-von Neumann computing architectures and other hardware-software co-designs based on CPU, GPU and FPGA have been extensively investigated to improve the computational efficiency~\cite{7551380,article1,eie}.

Among these innovative works, the brain-like neuromorphic computing appears as a promising solution: Deep networks are implemented by VLSI designs and high computing efficiency in speed and energy is obtained inherently by fulfilling data processing and communication in a single-chip~\cite{6055294}.
Neuromorphic designs with digital or analog computations have been reported not only in traditional CMOS technology but also in post-silicon devices such as spin devices and memristor~\cite{7167197,6055294,Sboev2017,Ankit:2017:RRE:3061639.3062311}. 
Contributed by the event-driven computation and digitized data communication, spiking neuromorphic computing (SNC) has been proved to be ultra-low-cost in design and energy and is highly attractive in deploying and executing deep networks.

\begin{figure}[b]
\centering
\vspace{-6pt}
\includegraphics[width=1\columnwidth]{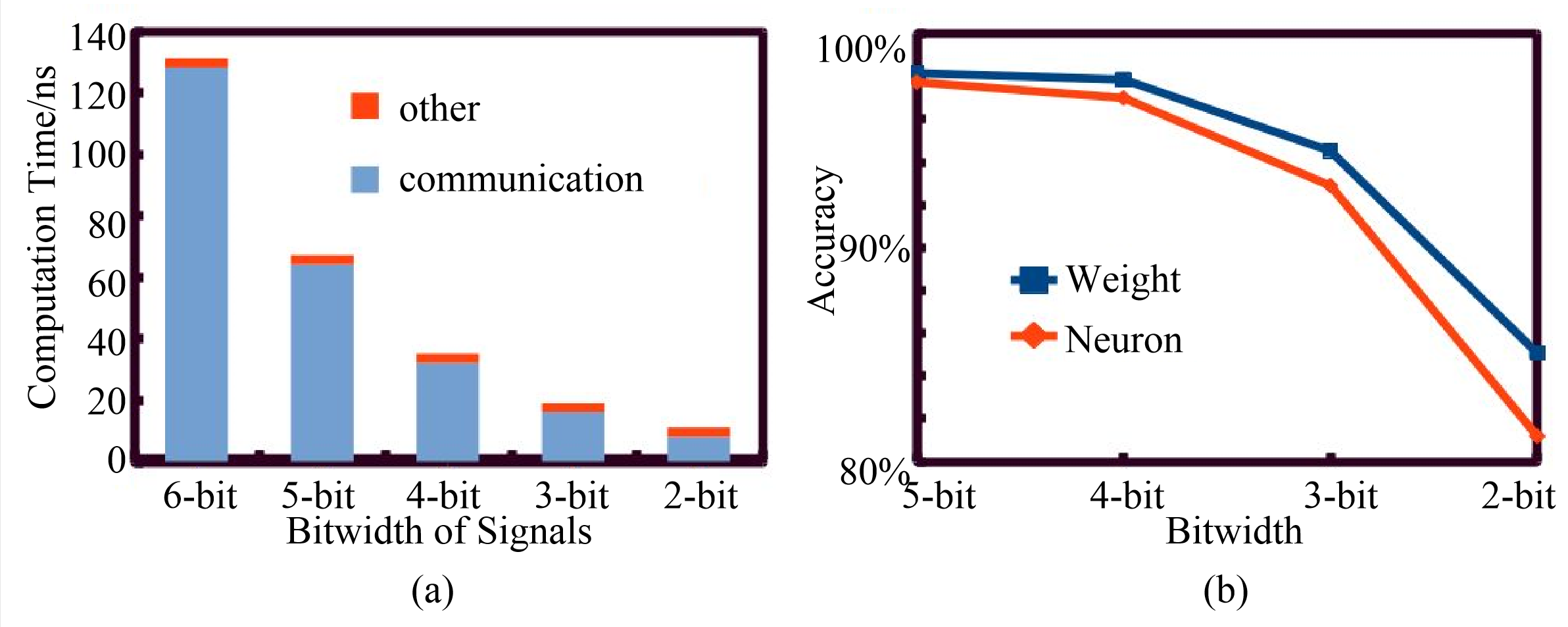}
\vspace{-18pt}
\caption{(a) Computation speed in different precision of neurons, (b) Accuracy loss caused by low-precision neurons and weights, respectively (evaluated on LeNet for MNIST dataset).}
\label{fig:system-scheme}
\end{figure}

The current system-level spiking designs mainly employ an off-line training methodology and the well-trained deep networks are deployed on the hardware system.
Nonetheless, one big challenge exists when performing the straightforward deployment, that is, obvious system accuracy loss induced by the constrained precision of \emph{synapses (or synaptic weights)} and \emph{neurons (or inter-layer signals)}.
For example, the IBM's TureNorth chip has five synaptic states (i.e. 0, $\pm$1, $\pm$2) and acceptable precision can only be achieved by assembling multiple synaptic layers with sacrificed design and energy cost~\cite{merolla2014}.   
Similarly, the synaptic weights in the memristor-based designs are usually represented by three or four bits data.
Although the memristor devices can afford continuous conductance states or 6-bit (64 levels) as was reported by HP Labs~\cite{Liu2017}, the heavy programming cost in speed and circuit design are not acceptable. 
In these SNC designs, the neuron signals are rate coded and the signal strength is represented by spike numbers in a time window in discrete values.
To get sufficient accuracy, the computation speed will be compensated, e.g. an 8-bit precision corresponds to 256 spikes and requires large time window for spike generation.

While solutions have been explored in previous works, the issue is still unsolved completely. In~\cite{7858418}, Wang \emph{et al.} proposed a one-level precision synapse and applied it on the memristor-based neuromorphic design, while the constrained precision of the neuron signals are unconsidered.
However, as is shown in Figure~\ref{fig:system-scheme}, computing speed of the spiking system is mainly constrained by data communication (i.e.  time required by spikes generation to guarantee good accuracy).
And compared to weight quantization,  discritizing the neurons results in larger accuracy loss.
In addition, realistic neuromorphic design of the proposed one-level synapse is challenging due to the various synaptic states distribution in different layers.
Recently, researchers tried a binary synapse and neurons deployment on TureNorth chip targeting high speed and low energy while retaining accuracy~\cite{article1}.
The  training rule is  similar to BinaryNet~\cite{NIPS2016_6573} and usually leads to obvious accuracy loss~\cite{imagenet}.

In this work, we focus on tackling the unacceptable accuracy loss caused by the low-precision spike neurons and synapses during deep networks deployment.
The 32-bit floating-point deep networks are transformed to data quantization-aware networks with fixed integer neurons and fix-point synapses.
The proposed networks can be applied to the emerging SNC universally. 
The memristor-based platform~\cite{7551380,7920854,7167197} is selected as our deployment example in this work.
Our target is retaining the high accuracy of deep networks in building dedicate hardware framework with high computing efficiency.
Our major contributions are summarized as follows:

\begin{itemize}
\item We transform the inter-layer signals of deep networks to be $M$-bit fixed integers in neural network training to mimic the discrete spike neurons in the SNC. 
These integral data in different layers are constrained to the same range and hence hardware implementation-friendly;
\vspace{-9pt}
\item We propose a weight clustering methodology to represent the synapses with $N$-bit fixed point data in a linear distribution. The best affordable states are obtained to improve system accuracy in low design cost;
\vspace{-9pt}
\item We deploy the proposed quantization-aware deep networks on the memristor-based SNC for performance evaluation.
The system accuracy on the state-of-the-art dataset such as MNIST and CIFAR-10 are measured. The speed, area, and energy are evaluated and compared with previous 8-bit fixed point design.
\end{itemize}

Our experimental results show that, when utilizing 4-bit integral neurons and fixed-point synapses and comparing with the ideal 32-bit floating point DNNs, our accuracy loss can be controlled within 0.02\% and 2.3\% on MNIST and CIFAR10.  
Compared with 8-bit fixed-point precision, our system can achieve more than 13.9$\times$ speedup, 89.1\% energy saving, and 30\% area saving.

\section{Preliminary}
\label{sec:pre}
\subsection{Related Works in DNNs Quantization}
\label{sec:pre:related}

Normally, the state-of-the-art deep networks are represented by 32-bit floating points.
Quantized DNNs have been explored in previous works to facilitate computation burden and hardware complexity while retain comparable accuracy.
Some earlier works focus on training DNNs with quantized weights and regardless of inter-layer signals~\cite{Darryl,haniclr2016}. 
For example, Lin \emph{et.al} trained the deep network efficiently with binary weights and quantized back propagation~\cite{Darryl}.

In recent works, implementation of DNNs with fixed-point synaptic weights and inter-layer signals are proposed.
Gysel \emph{et al.}~\cite{ICLR2016P} compressed DNNs into 8-bit dynamic fixed point values.
A fine-tuning was employed to recover the accuracy loss incurred by the weight quantization, however, the loss caused by the inter-layer signal quantization can not be recovered.
Adopting Gysel's quantization process, Tann \emph{et al.}~\cite{Tann} proposed DNNs with inter-layer signals in 8-bit dynamic fixed-point precision and weights in integer power-of-two values. 
Lin \emph{et al.}~\cite{Darryl} proposed to tune DNNs with fixed-point weights and inter-layer signals.
These works can achieve the target of improving computation efficiency in speed, hardware cost, and energy.
Unfortunately, they are not adaptive to the spiking neuromorphic systems in two reasons.
First, the 8-bit data utilization of inter-layer signals in the spiking systems will be extremely expensive in speed and hardware complexity.  
Second, the dynamic values varies greatly in the range for different layers and lead to large design complexity. 

Different with the above works, we implement quantized deep networks that be particularly feasible to the spiking neuromorphic systems: the proposed networks have fixed integer neurons and fixed point synapses and different layers have uniform values.


\begin{figure}[t]
\centering
\includegraphics[width=0.9\columnwidth]{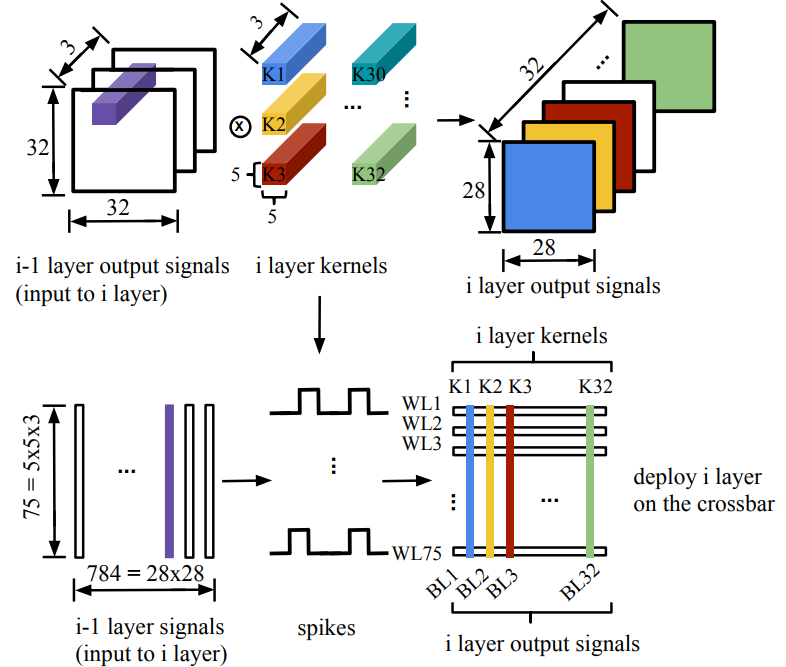}
\caption{Deploying a convolutional layer of DNN on crossbars}
\label{fig:mapping}
\end{figure}

\subsection{DNNs Deployment on SNC}
\label{sec:pre:deploy}

The memristor-based SNC platform is chosen to deploy DNNs in this work. 
Memristor is a two-terminal device in a MIM (metal-insulator-metal) structure that stores information by resistance states~\cite{6269919,HyunM,4585796}. 
Its high density crossbar structure and multiple states enable nature implementation of vector-matrix computation in a neural network, and thus  are extremely attractive to be leveraged in the emerging neuromorphic approaches~\cite{7551380,7920854,7167197}.

In the SNC implementation, weights in a neural network layer correspond to the memristor devices in a crossbar array.
Outputs of each layer are transformed to spikes and be fed into the next layer as inputs.
Fully connected layers in a DNN can be mapped on a crossbar directly~\cite{7167197}, while it is more complicate to implement a convolutional layer.
Figure~\ref{fig:mapping} depicts how to deploy a convolutional layer on the memristor-based SNC.
Filters in a convolutional layer is deployed to the crossbar column by column: $K^{i}_j$ that represents the $j^{th}$ convolution filter in the $i^{th}$ layer is mapped to the $j^{th}$ column of the crossbar, i.e. $BL_{j}$; 
the covolutional results of the filter $K^{i}_j$ is obtained at the end of $BL_{j}$.
Obviously, the $i^{th}$ layer with a filter number of $J^{i}$ requires crossbar with \emph{J} columns.
Consider the scenario that each 4-D filer in the $i^{th}$ layer has a size of ${s^i}\times{s^i}$ ($s^i$ is the scale of the filter) with a depth of ${d^i}$, the number of required rows will be ${s^i}\times{s^i}\times{d^i}\times{J^{i-1}}$~\cite{NIPS2012_4824}. 
Here, the ${d^i}$ is equals to $J^{i-1}$, which is the filer number of the $(i-1)^{th}$ layer.
Constrained by the realistic size of the memrisotr-based crossbar~\cite{GroupScissor}, multiple crossbar are utilized in parallel to compose a large layer. 
The crossbar numbers will be calculated as Equation~\ref{equ:number of crossbar}.
\begin{equation}\label{equ:number of crossbar}
L^{i} = \left \lceil \frac{J^{i}}{t}  \right \rceil \cdot \left \lceil \frac{s^i\times s^i\times J^{i-1}}{t}  \right \rceil.
\end{equation}
where 
$\left \lceil x\right \rceil=k,\ k\geq x\ \& \ x>(k-1) \ \& \ k\in \mathbb{Z}$, \emph{t} is the row or column size of a square crossbar.

\section{Design Methodology}
\label{sec:theoretical}

In this work, we aim to construct quantized DNNs with high accuracy, and hence obtain SNC with optimal accuracy, computation efficiency, and design cost.
Two major approaches are proposed--a \emph{Neuron Convergence} for fixed integer inter-layer signals and a \emph{Weight Clustering} for fixed-point weights. 
All network layers are executed and gain uniform values to minimize hardware design complexity.


\subsection{Neuron Convergence}
\label{sec:theoretical:neuron}
\begin{figure}[b]
\centering
\includegraphics[width=0.85\columnwidth]{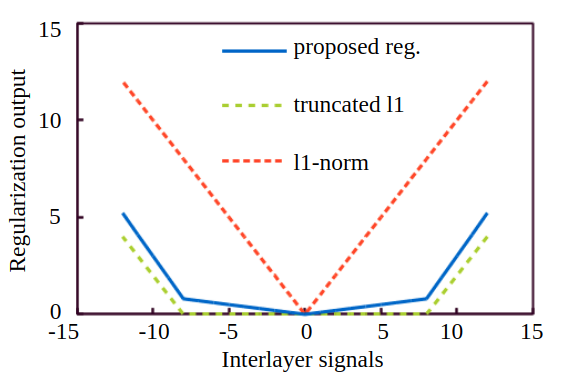}
\vspace{-12pt}
\caption{The forms of different regularization. Here, the bit width is set to be 2.}
\label{fig:reg}
\end{figure}

In this work, inter-layer signals are constrained to fixed integer in a dedicate range that is decided by target bit width.
The ranges are the same in all layers to achieve uniform values in networks and alleviate design complexity.
The fixed integer is adopted to mimic the discrete output spikes and the dedicate bit width is designed to decrease required spike numbers for high speed and low design cost.
Notably, quantizing inter-layer signals causes significant accuracy loss.
We propose a novel neuron regularization in neural network training to recover the loss.

As is indicated in Figure~\ref{fig:reg}, during neural networks training, $l1-norm$ regularization and truncated $l1-norm$ are usually utilized for weights sparsity and range restriction, respectively. 
In contrast, we propose a regularization term that can train neural networks with inter-layer signals not only sparse but also range-fixed. 
Particularly, the range in all the layers are uniform. 
The loss function in neural networks training are formulated as Equation~\ref{equ:new loss}.
\begin{equation}\label{equ:new loss}
E(W) = E_D(W) + \lambda \cdot R(W) + \sum_{i=1}^{L}(\lambda_i\cdot R_g(O^i)).
\end{equation}
Here, $W$ represents the weights in the DNN; $E_D(\cdot)$ is the loss term; $R(\cdot)$ is the normal regularization on weights.  
$R_g(\cdot)$ is the proposed regularization on each  inter-layer and its calculation in the $i^{th}$ layer can be represented as
$R_g(O^i)=\sum_{r=1}^{R^i}\sum_{c=1}^{C^i}\sum_{d=1}^{D^i}r_g(o^i_{r,c,d})$ (r, c, and d represent  row, column, and depth, separately).
The regularization of each inter-layer signal $r_g(\cdot)$ is calculated by Equation~\ref{equ:r}, as is described by the blue curve in Figure~\ref{fig:reg}.
\begin{equation}\label{equ:r}
r_g(o)=\left\{\begin{matrix}
\left | o\right |-2^{M-1} + \alpha \cdot \left | o \right |, & \left | o \right | \geq 2^{M-1}\ \\ 
\alpha \cdot \left | o \right |, & others 
\end{matrix}\right.
\end{equation}
$\alpha$ is set to be 0.1 empirically and $M$ is the targeting quantized bit width.

\begin{figure}[t]
\centering
\subfigure[none]{
\label{fig:bar1}
\includegraphics[width=0.2\textwidth]{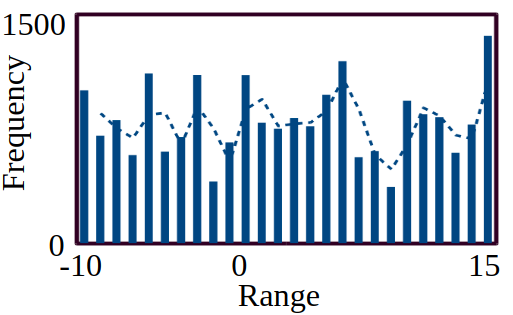}}
\subfigure[l1-norm]{
\label{fig:bar2}
\includegraphics[width=0.2\textwidth]{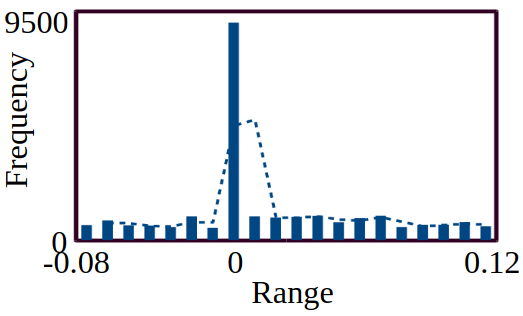}}\\
\subfigure[truncated l1]{
\label{fig:bar3}
\includegraphics[width=0.2\textwidth]{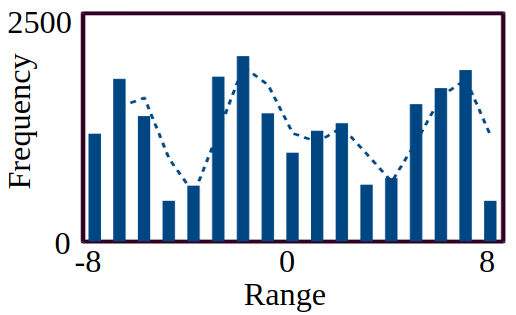}}
\subfigure[proposed reg]{
\label{fig:bar4}
\includegraphics[width=0.2\textwidth]{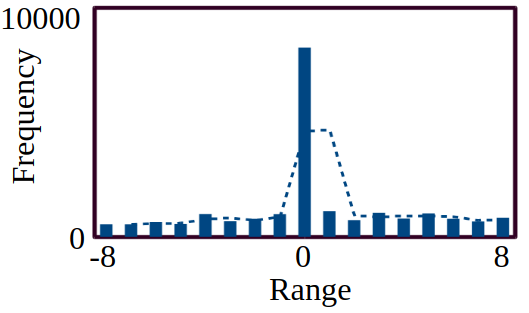}}
\vspace{-6pt}
\caption{Inter-layer signals distribution with applied regularization in four scenarios: none, $l1-norm$, truncated $l1-norm$ and the proposed. The $1^{st}$ hidden layer's outputs of LeNet on MNIST are shown as example and $M$ is set to be 4).}
\label{fig:Distributions} 
\end{figure}

Demonstrated by Figure~\ref{fig:Distributions}, our proposed regularization constrains the inter-layer signals in the objective range with sparse values successfully.
To obtain the target bit width, signals in the constrained range are then quantized to integer values. 
Note that the sparse and uniform range-fixed signal can greatly reduce the quantization loss inherently by minimizing the error transmitted from one layer to the next layer. 

More specifically, Equation~\ref{equ:error neural} illustrates the error between $o^i_{r,c,d}$ and its quantized value ${\hat{o}^i_{r,c,d}}$.
\begin{equation}\label{equ:error neural}
\bigtriangleup o^i_{r,c,d} = \sum_{d=1}^{d^{i-1}}\sum_{\theta=- \left \lfloor s^i/2 \right \rfloor }^{\left \lfloor s^i/2 \right \rfloor }\sum_{\theta=-\left \lfloor s^i/2 \right \rfloor }^{\left \lfloor s^i/2 \right \rfloor }\bigtriangleup o^{i-1}_{r+\theta,c+\theta,d} \cdot w^{i}_{j,\theta,\theta,d}.
\end{equation}
where $w^i_{j,\theta,\theta,d}$ represents the weight of the $j^{th}$ convolution filter in the $i^{th}$ layer, 
and $s^i$ is the size of the filter.
After the proposed training, the absolute value of the weights in the DNN should be extremely small as all the inter-layer signals are distributed in the same range.
Following Equation~\ref{equ:error neural}, the error $\bigtriangleup o^i_{r,c,d}$ is small and leads to mainly unchanged $\left \lceil o^i_{x,y,z} \right \rceil$.
As a result, the  error propagation is prevented and the quantized error is minimized.

Following the proposed DNNs training, the inter-layer signals are quantized to M-bit integer values with sparse property while retaining good accuracy.

\subsection{Weight Clustering}
\label{sec:theoretical:weight}
In implementing the  memristor-based SNC, floating points synaptic weights in a DNN are quantized to the available resistance states of the devices and result in accuracy drop.
We further propose a weight clustering to achieve fixed-point synaptic values in linear distribution that is hardware implementation friendly and can also reduce the accuracy loss.
Based on the inter-layer signals obtained in Sec.~\ref{sec:theoretical:neuron}, the accuracy loss generated by weight quantization can be represented by Equation~\ref{equ:error weight}.
\begin{equation}\label{equ:error weight}
\bigtriangleup o^i_{r,c,d} = \sum_{d=1}^{d^{i-1}}\sum_{\theta=- \left \lfloor s^i/2 \right \rfloor }^{\left \lfloor s^i/2 \right \rfloor }\sum_{\theta=-\left \lfloor s^i/2 \right \rfloor }^{\left \lfloor s^i/2 \right \rfloor } o^{i-1}_{r+\theta,c+\theta,d} \cdot \bigtriangleup w^{i}_{j,\theta,\theta,d}.
\end{equation}

Because of the sparse inter-layer signals  indicated in Figure~\ref{fig:Distributions}, the majority of $o^{i-1}_{r+\theta,c+\theta,d}$ are zero or close to zero. 
Similar to the explanation in Sec.~\ref{sec:theoretical:neuron}, the accuracy loss in the weight quantization is extremely small.
To further lower the loss, we train a cluster to minimize the error between the original weights and the quantized weights, as is depicted in Equation~\ref{equ:weight quantization}.
\begin{equation}\label{equ:weight quantization}
\centering
D^*=\underset{D}{argmin}\{\left \|  \frac{D}{2^N}-W\right \|^2\}
\end{equation}
where elements in $D$ belongs to $\{0,\ \pm 1,\ \pm 2,\ ...,\ \pm 2^{N-2},\ 2^{N-1}\}$, $N \geq log_2(\frac{max(\left | D\right |)}{max(\left | W\right |)})$, $W$ represents the weight matrix of a DNN, $\frac{D}{2^N}$ is the quantized matrix with fixed point, and \emph{N} is the target bit width of the weights.

The Equation~\ref{equ:weight quantization} is designed to find a matrix $\frac{D}{2^N}$, whose elements are $N$-bit fixed-point values with a linear distribution and best nearest the ideal floating point matrix in the DNN.
Here, we transform the weight quantization to an optimization problem that can be solved by k-nearest neighbors algorithm.





\section{Experiments}
\subsection{Experimental Setup}
\label{sec:exp}
Three different DNNs--Lenet, Alexnet, and Resnet are developed on Torch.
Neural networks model details and their ideal accuracy (Ideal Acc.) on MNIST and CIFAR10 without quantization are listed in Table~\ref{table:details}.
The quantized networks following our proposed method are implemented on the memristor-based SNC, and the hardware design methodology follows~\cite{7167197}.
Resistance states range of the memristor device is set to be [50K$\Omega$, 1M$\Omega$]~\cite{7167197}.
The crossbar size is set to be $32\times 32$, and required crossbar numbers of each network layer is calculated by the Equation~\ref{equ:number of crossbar} in Sec.~\ref{sec:pre:deploy}.
\begin{table}[!h] 
\vspace{-6pt}
\caption{Neural Network Models and Ideal Accuracy on MNIST and CIFAR10}  
\vspace{-6pt}
\begin{center}
\begin{tabular*}{3.4in}{llll}  
\hline  
Model & Lenet  & Alexnet & Resnet \\
Dataset & MNIST & CIFAR10 & CIFAR10\\
\hline  
Input Size  & $28\times 28\times1$ & $32\times32\times3$ & $32\times32\times3$\\  
Conv Layers  & $2(5\times 5)$ & $1(5\times 5), 4(3\times 3)$ & $17(3\times 3)$ \\  
FC Layers  &  2  & 3 & 1 \\
Weights &$7\times 10^3$& $3.4\times 10^5$ & $1.2\times 10^7$\\
Ideal Acc.  & 98.16\% & 85.35\% & 93.05\%\\ 
\hline  
\end{tabular*}
\end{center}
\label{table:details}
 \vspace{-12pt}
\end{table}




\subsection{Neuron Convergence on Inter-layer Signals Quantization}
\label{sec:exp:neuron}

The capability of \emph{Neuron Convergence} in recovering quantization accuracy loss is evaluated and results are listed in Table~\ref{table:neural quantization}.
In this experiment, the weights are ideal floating points without quantization.
Inter-layer signals of the Lenet, Alexnet, and Resnet are quantized to 5-bit, 4-bit, and 3-bit integer values by utilizing the proposed training and traditional training without the \emph{Neuron Convergence}.
As an example, accuracy of the two scenarios are represented by ``Lenet (w/)" and ``Lenet (w/o)"  in Table~\ref{table:neural quantization}. 
The accuracy recovered from traditional quantization by utilizing our proposed method is shown as ``Recovered Acc".
The computing accuracy of our proposed design is also compared with the idea accuracy and the accuracy loss is described in Table~\ref{table:neural quantization} as ``Acc. Drop".


\begin{table}[!h]
\caption{The Accuracy Measurement after Neuron Quantization with and without \emph{Neuron Convergence}}
\vspace{0pt}
\begin{center}
\begin{tabular*}{3.1in}{p{0.9in}p{0.6in}p{0.6in}p{0.6in}} 
\hline  
Model & 5-bit   & 4-bit & 3-bit \\
\hline
Lenet (w/o) & 97.74\%\  & 97\%\ & 92.9\%\ \\ 
Lenet (w/)  & 98.16\% & 98.15\% & 98.13\%\\  
Recovered Acc. & 0.42\%   & 1.15\% & 5.24\% \\
Acc. Drop&-0\%&-0.01\%&-0.03\%\\
\hline
Alexnet (w/o)& 82.51\% & 77.8\%& 67.83\%\\ 
Alexnet (w/)  & 85.2\% & 83.15\% & 82.1\%\\
Recovered Acc. & 2.69\%   & 4.95\% & 14.27\% \\
Acc. Drop&-0.15\%&-2.2\%&-3.25\%\\
\hline
Resnet (w/o)& 91.37\%& 75.72\% & 26.57\%\\ 
Resnet (w/)  & 92.5\% & 91.33\% & 88.95\%\\
Recovered Acc. & 1.13\%   & 15.61\% & 62.38\% \\
Acc. Drop&-0.55\%&-1.72\%&-4.1\%\\
\hline  
\end{tabular*}
\end{center}
\label{table:neural quantization}
\vspace{-12pt}
\end{table}

The results indicate quantizing inter-layer signals directly without the proposed \emph{Neuron Convergence} induces heavy accuracy loss, which is unacceptable. 
For example, the accuracy of  Alexnet and Resnet with 3-bit inter-layer signals on CIFAR10 drops to 67.83\% and 26.57\% from ideal accuracy 85.35\% and 93.05\%, respectively.
By utilizing our proposed method, the accuracy can be recovered to 82.1\% and 88.95\%.
The Lenet network on MNIST is robust and the 4-bit and 3-bit network has only 0.01\% and 0.03\% accuracy loss with our proposed training and discretize.
Our method can quantize the Alexnet and Resnet to 4-bit signals with 83.15\% and 91.33\% accuracy on CIFAR10. 
Compared with the ideal accuracy, the accuracy loss caused by the proposed 4-bit precision is only 2.2\% and 1.72\%.
The accuracy drop of the three networks in 5-bit signals are fully recovered (0\% on MNIST) or extremely small (0.15\% and 0.55\% on CIFAR10) after using our proposed method.

The above results demonstrate that our proposed \emph{Neuron Convergence} can recover the accuracy loss during signal quantization successfully. 
Neural networks with fixed integer signal and good accuracy are obtained. 
The best accuracy with 4-bit inter-layer signals on CIFAR10 can achieve 91.33\%, and the accuracy is 98.15\% on MNIST.


\subsection{Weight Clustering on Weights Quantization}
\label{sec:exp:weight}

\begin{table}[b]
\vspace{-12pt}
\caption{The Accuracy Measurement after Weights Quantization with and without \emph{Weight Clustering}}
\begin{center}
\begin{tabular*}{3.1in}{p{0.9in}p{0.6in}p{0.6in}p{0.6in}} 
\hline  
Model & 5-bit   & 4-bit & 3-bit \\
\hline
Lenet (w/o)& 98.16\% & 97.86\%& 94.52\%\\ 
Lenet (w/)  & 98.16\% & 98.1\% & 97.79\%\\  
Recovered Acc. & 0\%   & 0.24\% & 3.27\% \\
Acc. Drop &-0\%&-0.06\%&-0.37\%\\
\hline
Alexnet (w/o)& 83.02\% & 79.19\% & 75.33\%\\ 
Alexnet (w/)  & 85.26\% & 83.59\% & 82.92\%\\
Recovered Acc. & 2.28\%   & 4.4\% & 7.59\% \\
Acc. Drop &-0.05\%&-1.76\%&-2.43\%\\
\hline
Resnet (w/o)& 91\% & 77.12\% & 29\%\\ 
Resnet (w/)  & 92.8\% & 91\% & 88.1\%\\
Recovered Acc. & 1.8\%   & 12.88\% & 59.1\% \\
Acc. Drop &-0.25\%&-2.05\%&-4.95\%\\
\hline  
\end{tabular*}
\end{center}
\label{table:weight quantization}
\vspace{-12pt}
\end{table}

We also evaluate the performance of the proposed \emph{Weight Clustering} in recovering the accuracy loss caused by weights quantization.
Table~\ref{table:weight quantization} shows the experimental results with and without the proposed method.
Similarly, networks with 5-bit, 4-bit, and 3-bit fixed point weights are evaluated and the inter-layer signals are set to be ideal floating points without quantizaiton.
The results indicate that our proposed 4-bit Lenet, Alexnet, and Resnet can achieve 98.1\%, 83.69\%, and 91\% on MNIST and CIFAR10 with only 0.06\%, 1.76\%, and 2.05\% accuracy drop, comparing to the ideal accuracy.


\begin{table}[!h]  
	\caption{The Accuracy Measurement after Signals and Weights Quantization with and without our proposed method}  
	\begin{center}
	\begin{tabular*}{3.2in}{p{1in}p{0.6in}p{0.6in}p{0.6in}}  
		\hline
		Lenet 8-bit~\cite{ICLR2016P}& 98.16\%&&\\
		\hline  
		Model & 5-bit   & 4-bit & 3-bit \\
		\hline
		Lenet (w/o) &97.74\%&96.38\%&93.43\%\\
		Lenet (w/) &98.16\%&98.14\%&97.46\%\\
		Recovered Acc. &0.42\%&1.76\%&4.03\%\\
		Acc. Drop &-0\%&-0.02\%&-0.7\%\\
		\hline
		Alexnet 8-bit~\cite{ICLR2016P}& 84.5\%&&\\
		\hline
		Model & 5-bit   & 4-bit & 3-bit \\
		\hline
		Alexnet (w/o)& 81.8\%&76.16\%&69.7\%\\
		Alexnet (w/)& 84.47\%& 83.05\%&81.53\%\\
		Recovered Acc.& 2.67\%&6.89\%&11.83\%\\
		Acc. Drop& -0.88\%&-2.3\%&-3.82\%\\
		\hline
		Resnet 8-bit~\cite{ICLR2016P}& 91.75\%&&\\
		\hline
		Model & 5-bit   & 4-bit & 3-bit \\
		\hline
		Resnet (w/o)&  91.03\%&75.16\%&22.18\%\\
		Resnet (w/)& 91.48\%&90.33\%&87.71\%\\
		Recovered Acc.& 0.45\%&15.17\%&65.53\%\\
		Acc. Drop& -1.57\%&-2.72\%&-5.34\%\\
		\hline
	\end{tabular*}
	\end{center}
	\label{table:performance}
	\vspace{-12pt}
\end{table}

\subsection{Neuron Convergence and Weights Clustering on Data Quantization}
\label{sec:exp:weight and neuron}

\begin{table*}[!h]
	\caption{Memristor-bsed SNC System Evaluation and Comparison}  
	\vspace{3pt}
	\begin{center}
	\begin{tabular*}{6in}{l|c|c|c|c|c|c|c}  
		\hline
		\multirow{2}{*}{model}&\multirow{2}{*}{Layer Num.}&$Speed$&\multirow{2}{*}{Speedup}&$Energy$&Energy&$Area$&Area \\
		&&$(MHz)$&&$(\emph{uJ})$&Saving&$(mm^2)$&Saving\\
		\hline
		Lenet 8-bit~\cite{ICLR2016P}&4&$0.64$&-&$4.7$&-&$1.48$&-\\
		Lenet 4-bit in this work&4&$8.93$&13.9x&$0.57$&87.9\%&$1.04$&29.7\%\\
		Lenet 3-bit in this work&4&$15.63$&24.4x&$0.27$&94.3\%&$0.93$&37.2\%\\
		\hline
		Alexnet 8-bit~\cite{ICLR2016P}&8&$0.27$&-&$337.0$&-&$34.3$&-\\
		Alexnet 4-bit in this work&8&$2.66$&9.8x &$36.9$&89.1\%&$24.0$&30\%\\
		Alexnet 3-bit in this work&8&$3.79$&11.8x&$26.3$&92.2\%&$21.4$&37.6\%\\
		\hline
		Resnet 8-bit~\cite{ICLR2016P}&18&$0.11$&-&$19200$&-&$937.3$&-\\
		Resnet 4-bit in this work&18&$1.38$&12.5x&$1500$&92.2\%&$656.2$&30\%\\
		Resnet 3-bit in this work&18&$2.20$&20x&$935$&95\%&$585.9$&37.5\%\\
		\hline
	\end{tabular*}
	\end{center}
	\label{table:final result}
	\vspace{-6pt}
\end{table*}

In this experiment, the proposed \emph{Neuron Convergence} and \emph{Weights Clustering} are applied together in the three neural networks for overall performance evaluation.
Through the proposed method, the inter-layer signals and the weights are quantized to fixed integer values and fixed-point values in 5-bit, 4-bit, and 3-bit, respectively.
The accuracy of the networks with and without the proposed method is depicted and compared in Table~\ref{table:performance}.
Similar to Sec.~\ref{sec:exp:neuron} and Sec.~\ref{sec:exp:weight}, the ``Recovered Acc." indicates accuracy recovery ability of our proposed method and ``Acc. Drop" is the accuracy loss compared with the ideal accuracy.
Besides compared with the ideal accuracy in Table~\ref{table:details}, we also include the accuracy of the 8-bit dynamic fixed point neural networks in~\cite{ICLR2016P} for comparison.

Compard with the 8-bit dynamic fixed point networks in~\cite{ICLR2016P}, our proposed networks with 5-bit integer inter-layer signal and fixed-point weights can gain almost the same accuracy: same accuracy of Lenet on MNIST, only 0.03\% drop of Alexnet on CIFAR10, and 0.72\% drop of Resnet on CIFAR10.
Our proposed 4-bit Lenet, Alexnet, and Resnet can achieve accuracy of 98.14\%, 83.05\%, and 90.33\% on MNIST and CIFAR10 with only 0.06\%, 1.76\%, and 2.05\% accuracy loss compared  with the ideal accuracy.
Even with 3-bit data representation, our method can achieve 97.46\% on MNIST and 87.71\% on CIFAR10.

Based on the above discussions, it is proved that our proposed method can represent DNNs using 4-bit or even 3-bit data representation in the inter-layer signals and weights while keeping good accuracy. 



\subsection{Improvement on Computation Efficiency }

In the SNC system implementation, computation result of one DNNs layer is transformed to spikes by integrate-and-fire circuits (IFCs) to generate digitized outputs through counters~\cite{7167197,6055294}.
Therefore, a reduced bit width of signals between layers corresponds to less required spike numbers and thus improved speed, design cost, and energy efficiency.
Weights quantization also helps to improve computing efficiency and reduce hardware design complexity by decreasing the utilization of synaptic crossbar and programming cost. 

In this work, the benefit of our proposed DNNs with \emph{M} bit fixed integer inter-layer signals and \emph{N} bit fixed-point weights on improving computation efficiency is evaluated on the memristor-based SNC.
Based on the results in Sec.~\ref{sec:exp:weight and neuron}, two scenarios with (\emph{M}, \emph{N}) is (4, 4) and  (3, 3) are implemented and analyzed.
The 8-bit dynamic fixed-point in~\cite{ICLR2016P} is also implemented for comparison.
In the memristor-based SNC, each computation unit (i.e. neural network layer) includes four components: wordline (WL) drivers to generate robust input signals, memristor-based crossbars to complete the matrix computation, IFCs to convert the current results from the crossbar to spikes, and counters to generate digitized output of each layer.
The speed, energy, and area are obtained from circuits simulation on IBM 130nm technology and the simulation parameter configuration is based on~\cite{7167197}.

The results in  Table~\ref{table:final result} show that our proposed method can achieve significant computation efficiency improvement compared with the previous 8-bit dynamic fixed point DNNs.
Our systems have more than 9.8$\times$ speed up, 89.1\% energy saving and 29.7\% area saving. 

\section{Conclusions}

DNNs quantization in implementing the spiking neuromorphic computing (SNC) is important for acceptable design complexity and computational efficiency. 
However, directly weights and inter-layer signal quantization cause heavy accuracy loss.
In this work, we propose data quantization-aware DNNs with a neuron convergence and a weight clustering method to recover the accuracy loss in neural network quantization.
The obtained fixed integer signals and fixed-point weights particularly benefit the SNC in design cost and computation efficiency.
We carefully deploy the quantized DNNs on the memristor-based SNC to study the system efficiency improvement that can be achieved by the proposed method.
The system accuracy and performance is evaluated in three networks--Lenet, Alexnet, and Resnet on MNIST and CIFAR10 and compared with the ideal DNNs and the previous 8-bit dynamic fixed-point DNNs.
The results indicate that the design can achieve 98.14\% and 90.33\% accuracy on MNIST and CIFAR10 with 4-bit data representation, which is only 0.02\% and 2.72\% lower than the ideal DNNs.
Compared with the  8-bit dynamic fixed point framework, the proposed design demonstrates more than 9.8$\times$ speedup, 89.1\% energy saving, and 30\% area saving.

\vspace{-3pt}
\section{Acknowledgments}

This work is supported in part by AFRL ICA2017-UP-017. We would like to thank NVIDIA Corporation for their generous GPU donation. Any opinions, findings, and conclusions or recommendations expressed in this material are those of authors and do not necessarily reflect the views of AFRL or its contractors.

\bibliographystyle{ieeetr}
\small
\bibliography{ref} 

\end{document}